\documentclass[10pt,twocolumn,letterpaper]{article}

\usepackage[pagenumbers]{cvpr} 

\usepackage{graphicx}
\usepackage{amsmath}
\usepackage{amssymb}
\usepackage{booktabs}
\usepackage{color}
\usepackage{xcolor}
\usepackage{tabularx}

\usepackage{makecell}
\usepackage{multirow}
\usepackage{arydshln}

\usepackage{amsmath,amsfonts,bm}

\def\1{\bm{1}}

\def\vzero{{\bm{0}}}

\def\vc{{\bm{c}}}

\def\vn{{\bm{n}}}

\def\vr{{\bm{r}}}

\def\vt{{\bm{t}}}

\def\vv{{\bm{v}}}

\def\vx{{\bm{x}}}

\def\mI{{\bm{I}}}

\def\mK{{\bm{K}}}

\def\mP{{\bm{P}}}

\def\mR{{\bm{R}}}
\def\mS{{\bm{S}}}

\def\mU{{\bm{U}}}
\def\mV{{\bm{V}}}

\DeclareMathAlphabet{\mathsfit}{\encodingdefault}{\sfdefault}{m}{sl}
\SetMathAlphabet{\mathsfit}{bold}{\encodingdefault}{\sfdefault}{bx}{n}

\def\gL{{\mathcal{L}}}

\newcommand{\R}{\mathbb{R}}

\DeclareMathOperator{\sign}{sign}

\newcommand{\transposeSign}{{\!\top\!}}
\newcommand{\tr}{\transposeSign}

\usepackage[pagebackref,breaklinks,colorlinks]{hyperref}

\usepackage[capitalize]{cleveref}
\crefname{section}{Sec.}{Secs.}
\Crefname{section}{Section}{Sections}
\Crefname{table}{Table}{Tables}
\crefname{table}{Tab.}{Tabs.}
\Crefname{figure}{Figure}{Figures}
\crefname{figure}{Fig.}{Figs.}

 \renewcommand{\paragraph}[1]{\smallskip\noindent{\bf{#1}}}
 
\usepackage[symbol]{footmisc}

\usepackage[belowskip=4pt,aboveskip=4pt]{caption}

\begin{document}

\title{Input-level Inductive Biases for 3D Reconstruction}

\author{
Wang Yifan\textsuperscript{1}\footnotemark
\hspace{1.0em}
Carl Doersch\textsuperscript{2}\hspace{1.0em}
Relja Arandjelović\textsuperscript{2}\hspace{1.0em}
João Carreira\textsuperscript{2}\hspace{1.0em}
Andrew Zisserman\textsuperscript{2,3}\hspace{1.0em}
\\\\
\textsuperscript{1}ETH Zurich \hspace{1.0em} \textsuperscript{2}DeepMind \hspace{1.0em}
\textsuperscript{3}VGG, Department of Engineering Science, University of Oxford
\vspace{-8pt}
}

\twocolumn[{%
\renewcommand\twocolumn[1][]{#1}%
\vspace{-8ex}
\maketitle
\begin{center}
	\includegraphics[width=\linewidth]{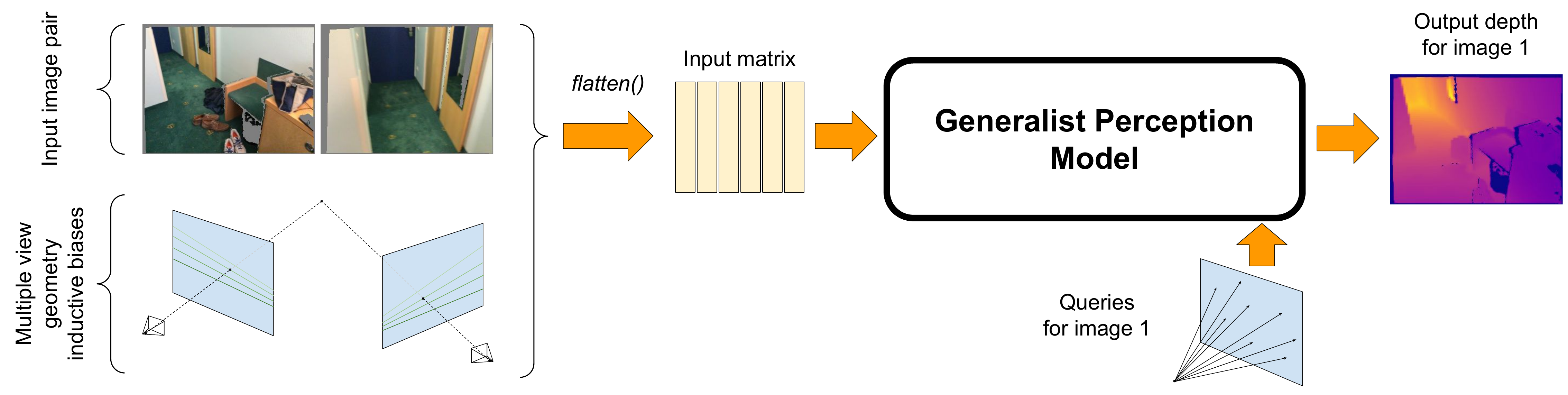}
\captionof{figure}{
{\bf Input-level inductive biases.}
	We explore 3D reconstruction using a generalist perception model, the recent Perceiver IO~\cite{jaegle2021perceiverIO} which ingests a matrix of unordered and flattened inputs (\eg pixels). The model is interrogated using a query matrix and generates an output for every query -- in this paper the outputs are depth values for all pixels of the input image pair.
	We incorporate inductive biases useful for multiple view geometry into this generalist model without having to touch its architecture, by instead encoding them directly as additional inputs.
}\label{fig:teaser}
\end{center}
}]

\maketitle

\begin{abstract}
Much of the recent progress in 3D vision has been driven by the development of specialized architectures that incorporate geometrical inductive biases. In this paper we tackle 3D reconstruction using a domain agnostic architecture and study how to inject the same type of inductive biases directly as extra inputs to the model. This approach makes it possible to apply existing general models, such as Perceivers, on this rich domain, without the need for architectural changes, while simultaneously maintaining data efficiency of bespoke models. In particular we study how to encode cameras, projective ray incidence and epipolar geometry as model inputs, and demonstrate competitive multi-view depth estimation performance on multiple benchmarks.
\end{abstract}

\footnotetext{$^{*}$Work done during internship at DeepMind.}

\section{Introduction}
\label{sec:intro}

The focus of modern computer vision research is, to a large extent, to identify good architectures for each task of interest. There are many tasks of interest, ranging from classical ones such as optical flow~\cite{horn1981determining}, to highly specialized (yet arguably important) ones such as recognizing equine action units~\cite{li2021automated}. Creating dedicated models for every possible task naturally results in a sprawling catalog of architectures.

Eventually it seems desirable to build a more general visual system that can deal with most perceptual problems. To get there, one option is to combine state-of-the-art systems for all of those problems, but this would be complex, inelegant and not scalable. Another option is to employ models without much customization or inductive biases for any particular task, but these models will by definition be less data-efficient and hence less accurate than specialized ones given a fixed data budget. 

In this paper we explore the single-general-model route. We ask the following question: can the lack of architecture-level inductive biases be replaced by extra inputs which encode our knowledge about the problem structure? In other words, can we feed those priors as inputs rather than hardwire them into the model architecture (\cref{fig:teaser}), like a loadable software solution instead of a more rigid hardware solution. As the general model we employ the recently published Perceiver IO~\cite{jaegle2021perceiverIO} and as domain we focus on multiview geometry and 3D reconstruction, an area of computer vision where architectural specialization is particularly exuberant~\cite{rocco2018neighbourhood,zhou2018deeptam,zhang2019ga,melekhov2019dgc,kusupati2020normal,huang2018deepmvs,im2019dpsnet}.  

Our main contribution is in mapping out and evaluating some of the options for expressing priors for 3D reconstruction as input features, in particular in the setting of depth estimation from stereo image pairs. We consider concepts in multiview geometry such as camera viewpoint, light ray direction and epipolar constraints. Similar to the prior work we compare with~\cite{im2019dpsnet,huang2018deepmvs,kusupati2020normal,ummenhofer2017demon}, we assume ground truth cameras are given, but they could in principle be computed by the model as well and passed back as inputs recurrently.

We experiment on multiple datasets---ScanNet~\cite{dai2017scannet}, SUN3D~\cite{xiao2013sun3d}, RGBD-SLAM~\cite{sturm2012benchmark} and Scenes11~\cite{ummenhofer2017demon}---and present results that are comparable or better to those obtained by state-of-the-art specialized architectures on all of them. This is achieved without using cost volumes, warping layers, etc., and in fact (proudly) without introducing any architectural innovation. Instead, we propose powerful input-level 3D inductive biases that substantially improve data efficiency.
This paper reflects a new avenue for problem solving in computer vision, in which domain knowledge is valued but applied in a flexible manner, as additional model inputs.

\section{Related Work}
Our work is part of a long trend in computer vision of simplifying and unifying architectures.  It was noted a decade ago that big data along with simple architectures are ``unreasonably effective''~\cite{halevy2009unreasonable} at solving many perception problems, and subsequent progress has only reinforced this~\cite{sutton2019bitter}.  Computer vision has moved from architectures like ConvNets, which are highly general image processors~\cite{krizhevsky2012imagenet}, to methods that are based on Transformers~\cite{vaswani2017attention} such as ViT~\cite{dosovitskiy2020image} and Perceivers~\cite{jaegle2021perceiver,jaegle2021perceiverIO}, where the underlying Transformer can be equally effective across multiple domains like sound and language. Unifying architectures is useful because architectural improvements can be propagated across tasks and domains trivially. It also enables sharing and transferring information across modalities and tasks~\cite{yosinski2014transferable,sharif2014cnn}, which is critical for tasks with little data.

However, seeking general-purpose architectures does not mean we should discard insights about geometry when solving a geometric problem.  Decomposing the problem into feature matching and triangulation was an early component of stereo systems~\cite{ohta1985stereo,hirschmuller2007stereo}. More recent systems have relied on learning, especially for learning descriptors which are compared across images to find correspondence, either by directly searching for matches across images~\cite{chen2015deep,mayer2016large,liang2018learning,xu2020aanet} or by computing 4D correlation volumes~\cite{chang2018pyramid,kendall2017end,zhang2019ga,yang2019hierarchical,cheng2020hierarchical,yang2020cost,gu2020cascade}, or a combination~\cite{guo2019group}; scaling these methods can be problematic as the number of considered matches grows. 
Several recent works~\cite{he2020epipolar,yang2021mvs2d,li2021revisiting} inferred correspondences by aggregating sample points along the epipolar line with a transformer; however matches are still represented and sampled explicitly.
Similar to our work, Cam-Convs~\cite{facil2019cam} leveraged input-level geometric priors (camera intrinsics) for more robust single-view depth estimation under variable camera. 
Our work considers a more general application - multi-view depth estimation, where we include also the camera relative poses and the epipolar embedding.

The broader field of correspondence learning has a variety of approaches for integrating global and local inference. Early approaches to deep optical flow and correspondence estimation~\cite{dosovitskiy2015flownet,detone2016deep} used direct regression, as our approach does, but later works found correlations and cost volumes more effective~\cite{dosovitskiy2015flownet,sun2018pwc,teed2020raft}.  
Perceiver IO~\cite{jaegle2021perceiverIO}, however, shows strong flow performance with direct regression.  
Transformers have also contributed to improvements in more general scene correspondence~\cite{jiang2021cotr,sun2021loftr,wei2021fast}, and even using learned correspondence to improve few-shot learning~\cite{doersch2020crosstransformers}, though these transformers are still applied on feature grids with relatively complex mechanisms to represent correspondence explicitly. The grids are taken from prior work on correspondence that uses deep learning, where explicit pairwise comparisons and cost volumes are a staple of top-performing methods~\cite{rocco2020efficient,rocco2018neighbourhood,choy2020deep,melekhov2019dgc,min2019hyperpixel,sarlin2020superglue,yi2018learning}.

Our work also sits in the broader field of deep learning for 3D reconstruction, where there have been a wide variety of proposals for representing 3D inductive biases.  Early works like DeepTAM~\cite{zhou2018deeptam} emphasize the importance of representing per-image depth maps and rays.  More recent works have made use of deep implicit models to represent 3D~\cite{park2019deepsdf,mescheder2019occupancy,chen2019learning}, introducing the idea that deep representations should be queried with points.  
While this work has been extended to more complex scenes in NeRF~\cite{mildenhall2020nerf} and its many derivatives, these typically require many images of the same scene and an expensive offline training process.  Online methods typically rely on more explicit but expensive 3D representations like voxel grids~\cite{zhang2021learning,sun2021neuralrecon,ji2017surfacenet,murez2020atlas}.  Particularly relevant is TransformerFusion~\cite{bovzivc2021transformerfusion}, which uses Transformers to attend from its voxel grid representation to the input images, although this approach still suffers from problems with memory and resolution due to the voxel grid.

\subsection{Review of Perceiver IO}
\label{sec:perceiver}

For the general perception model we use Perceiver IO~\cite{jaegle2021perceiverIO},
and we briefly review it here.
The model is based on Transformers~\cite{vaswani2017attention} in that it treats its input as a simple series of tokens, and attention is the main workhorse. First, cross-attention is performed between the input tokens and a fixed-size set of internal vectors (`latents'), thus obtaining a compressed representation of the input. Then, a series of self-attentions is performed within the latents, enabling this architecture to scale well to large inputs (\eg high resolution images) and to stack many layers without hitting memory issues, since there are much fewer latents than input tokens.  The final step is another cross-attention, this time between a set of externally specified `queries' and the latents, which produces an output array of desired size (one element for each query). Queries are typically some encoding of pixel position and are quite dense (\eg one per pixel).
The architecture achieves strong results on a large variety of tasks and domains, such as image classification, optical flow, natural language understanding, and StarCraft II,
making it a natural fit for the general perception model used in this work (\cref{fig:teaser}).

\section{Featurizing Multiple View Geometry}

\begin{figure}[t!]
	\centering
	\includegraphics[width=\linewidth]{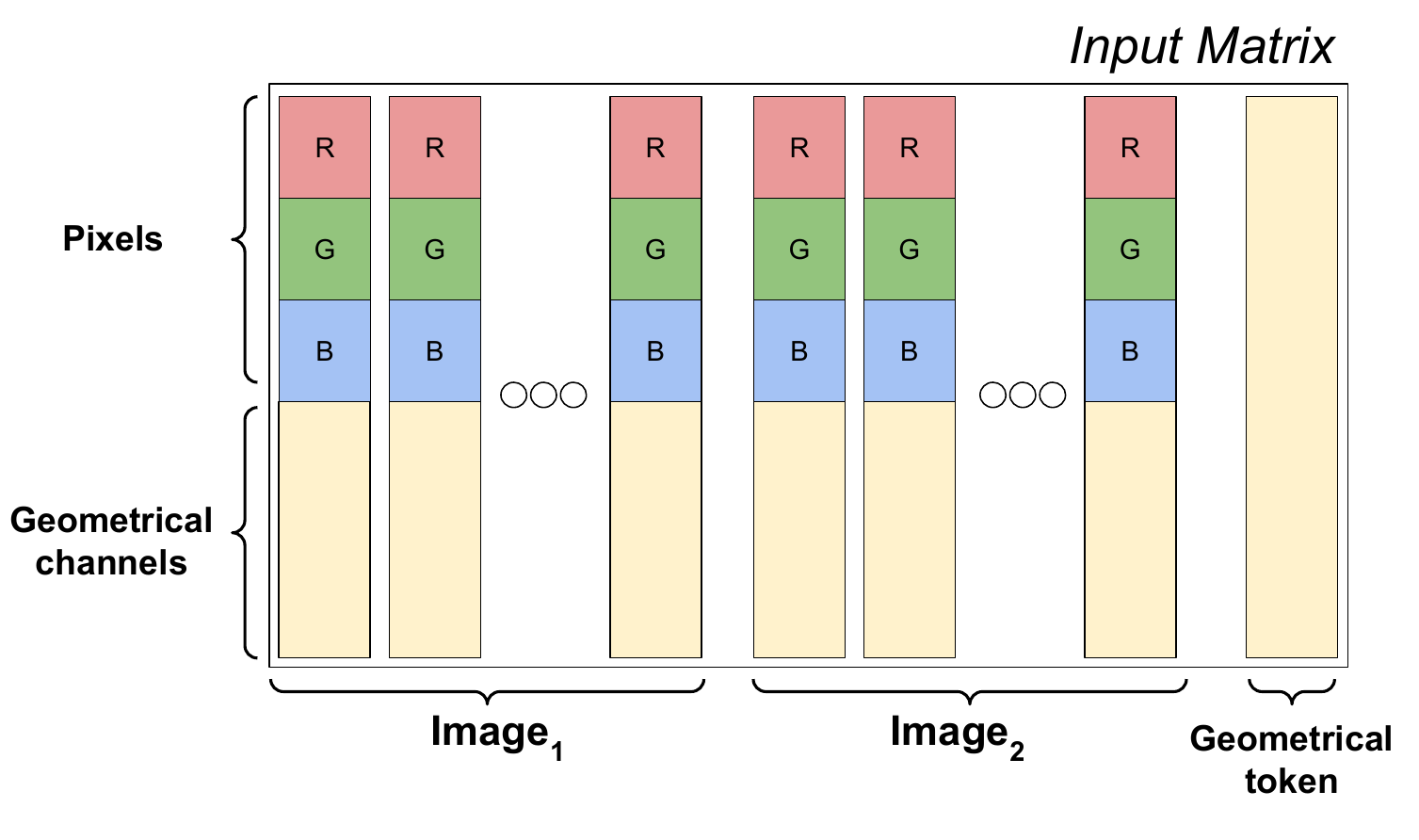}\vspace{-1ex}
	\caption{{\bf Geometrical Embeddings.} The input to a Perceiver model is a matrix. Instead of vanilla positional embeddings, we introduce geometrical embeddings that encode inductive biases from multiple view geometry. We form the input matrix by concatenating pixel values with these embeddings: as extra per-pixel channels and/or as extra tokens.} 
	\label{fig:input_matrix}\vspace{-3ex}
\end{figure}

In this section, we demonstrate how to inject geometric inductive biases into a general perception model, Perceiver IO~\cite{jaegle2021perceiverIO} (\cref{sec:perceiver}), without changing its architecture.
We consider the case of 3D reconstruction from an image pair -- the inputs are pixels and calibrated cameras, and the output is depth at each pixel.

If we follow prior work, such as the optical flow network from Perceiver IO~\cite{jaegle2021perceiverIO}, then we can treat each pixel (or, more generally, each vector in a feature grid) as an input element.  We then tag each pixel with an encoding for its position within the grid as input, and potentially with an additional tag to indicate which of the two input images the pixel belongs to.  The output could be processed similarly: we use the same tagged pixels (or features) as queries in order to get a depth value for each pixel.

In practice, however, we expect this approach to overfit given the relatively small datasets that are available for training geometric inference.  A high-capacity model can easily memorize the depth for each image, rather than learning a procedure which matches features across images and performs triangulation in a way that can generalize to unfamiliar scenes.

Our hypothesis is that we can create a more data-efficient learning algorithm by simply providing the Perceiver IO with information that describes the geometry as input.  In the ideal case, Perceiver IO can learn to use this information correctly without the computational pipeline being prescribed by a complex, restrictive architecture.

In particular, we explore providing information that lets the network more easily 1) represent 3D space to allow triangulation, and 2) find correspondences, which are the two main components of any general stereo system.  Towards 1, we explore providing camera information in the form of encoded camera matrices, as well as the encodings of the rays at every pixel.  Towards 2, we encode epipolar planes for each pixel, which tells the network which pixels might be in correspondence.  Our main contribution in this work is to show that together, these geometric quantities can improve the inferred 3D geometry without any changes to the network architecture.

The geometric information is provided as input to the network. We explore two main ways to operationalize this (\cref{fig:input_matrix}): 1) by fusing the information with all input elements via concatenating it along the channel dimension, and 2) by expanding the input set with additional `geometrical' tokens.

\subsection{Featurizing Cameras}\label{sec:cameras}
A camera is one of the most important components for multiview geometry, providing the necessary information to perform triangulation~\cite{Hartley04c}.
We assume the commonly used pin-hole camera model parameterized with the intrinsic parameters \(\mK\in\R^{3\times 3}\), which define the transformation from camera coordinates to image coordinates, and extrinsic parameters \(\left[\mR\in\R^{3\times 3}, \vt\in\R^{3}\right]\), defining the 6-DOF camera pose transformation from the world coordinates to the camera coordinates.
In practice the intrinsic parameters can be obtained by off-the-shelf calibration methods~\cite{zhang2000flexible} and the extrinsic parameters can be estimated using structure-from-motion algorithms such as COLMAP~\cite{schoenberger2016sfm}.

Next, we consider two alternatives to encode the camera parameters, the first one is based on constructing viewing rays connecting the camera with each pixel, and the second is directly providing the projection matrix that maps the 3D world coordinates to the 2D pixel coordinates.

\begin{figure}[t!]
	\centering
	\includegraphics[width=\linewidth]{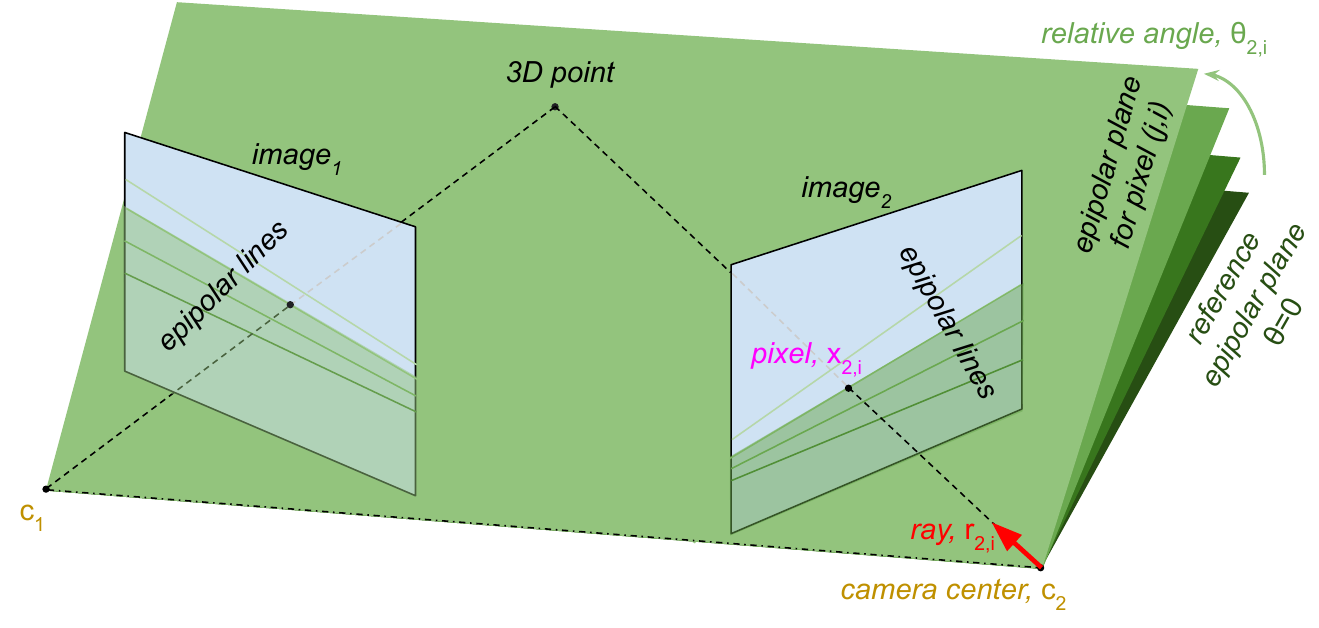}
	\caption{Geometric entities used to compute the geometrical embeddings that are passed as inputs to the perception module.
	For clarity, only entities related to one of the images are labeled.}
	\label{fig:geometry}\vspace{-3ex}
\end{figure}

\paragraph{Option 1: Rays and camera center.}
Let \(\vx_{j,i} \in\R^{2}\) be the image coordinate of pixel \(i\) in image \(j\). It can be uniquely represented in the 3D space using the viewing ray, which can be further parameterized using the camera center, \(\vc_{j}\in\R^{3}\), and the unit-length ray direction, \(\vr_{j,i}\in\R^{3}\) (\cref{fig:geometry}).
The projection matrix for the camera $j$ is a $3\times4$ matrix $\mP_{j} = \mK_{j} \left[\mR_j | \vt_j \right]$. 
In homogeneous coordinates, the camera center $ \tilde{\vc}_j = \left[\vc_j, 1\right]^\tr$, satisfies $\mP_j\tilde{\vc_j} = \vzero $.
Writing the projection matrix as $\mP_j = \left[ \mK_j\mR_j | \mK_j\vt_j \right] $, the camera center in the world coordinate system is
\begin{equation}
\vc_{j} = -(\mK_j\mR_j)^{-1}\mK_j\vt_j=-\mR_j^{-1}\vt_j. \label{eq:camera_center}
\end{equation}
The unnormalized viewing ray direction can be computed as
\begin{equation}
    \bar{\vr}_{j,i} = \left(\mK_j\mR_j\right)^{-1}\begin{bmatrix}\vx_{j,i}\\1\end{bmatrix}, \label{eq:camera_rays}
\end{equation}
since $\mP_j \left[\bar{\vr}_{j,i}, 0\right]^\tr = \left[\vx_{j,i}, 1\right]^\tr$, which we normalize to unit length to obtain $\vr_{j,i}$.

Instead of providing $\vc_{j}$ and $\vr_{j,i}$ to the network in their raw form as 3-D vectors, we embed them to higher dimensional Fourier features, as it was shown empirically that this higher-dimensional encoding is better suited for further processing by neural networks~\cite{mildenhall2020nerf,vaswani2017attention}.
This is done by applying element-wise mapping $x\mapsto\left[x, \sin(f_1\pi x), \cos(f_1\pi x), \cdots, \sin(f_K\pi x), \cos(f_K\pi x)\right]$, where $K$ is the number of Fourier bands, and $f_k$ is equally spaced between 1 and $\frac{\mu}{2}$, with $\mu$ corresponding to the sampling rate.
The sampling rate $\mu$ and number of bands $K$ are hyperparameters which can be set separately for $\vc_{j}$ and $\vr_{j,i}$.
As a result, we obtain $6K_{c}+3$ and $6K_{r}+3$ Fourier features for $\vc_{j}$ and $\vr_{j,i}$ respectively.

\paragraph{Option 2: Pixel coordinates and projection matrix.}
Alternatively, since the 3D position of each pixel can be determined up to an unknown depth solely using the projection matrix $\mP_j$, we can also uniquely embed each pixel directly with $\mP_j$ and the pixel coordinate $\vx_{j,i}$.
To this end, we flatten $\mP_j$ to a 12-dimensional vector, then we map this 12-D vector as well as the 2-D $\vx_{j,i}$ again to Fourier features using $K_{\textrm{matrix}}$ and $K_{x}$ bands, and $\mu_{\textrm{matrix}}$ and $\mu_{x}$ sampling rates, respectively.
The resulting $24K_{\textrm{matrix}}+12$ and $4K_{x}+2$ vectors uniquely determine the geometry for a given pixel.

\paragraph{Injecting the camera information into the input of Perceiver IO.}
The above geometric embeddings contain all the necessary information for the network to triangulate the pixels. We now consider how to provide this information to the general perception model.
Notice that in either aforementioned options, there is a camera-specific part which is identical to all pixels in the a given image, namely $\vc_{j}$ and  $\mP_j$,
and a pixel-specific part that is unique to each pixel, namely $\vr_{j,i}$ and $\vx_{j,i}$.
The pixel-specific part is most naturally incorporated by concatenating it with the pixel's RGB values along the channel dimension (`geometrical channels' in~\cref{fig:input_matrix}). There are two ways of assembling the camera-specific part -- again as \emph{geometrical channels}, or as additional \emph{geometrical tokens}.

The first way consists of simply duplicating the camera-specific embedding for all the pixels of the corresponding image and again concatenating it along the channel dimension as \emph{geometrical channels}.
This results in a total of $2\times H\times W$ inputs of $\left(D_{\textrm{rgb}} + D_{\textrm{pix}} + D_{\textrm{cam}}\right)$ dimensions, where $\left(H, W\right)$ is the image dimension and $D_{\textrm{rgb}}$, $D_{\textrm{pix}}$ and $D_{\textrm{cam}}$ are the total dimensions of the RGB-based inputs, pixel-specific and camera-specific geometry embeddings respectively.

Alternatively, we can treat the camera-specific embedding as a separate input \emph{geometric token}, alongside the per-pixel inputs, yielding a total of $2\times\left(H\times W+1\right)$ input tokens.
In order to indicate which image is a pixel associated with, we append an additional image indicator embedding to the per-pixel tokens, that is unique per image and shared among all the pixels in the same image.
In our experiment, we encode this image indicator as a $D_{\textrm{ind}}$-dimensional vector using either a Fourier mapping of the image index (0/1), or as a learnable parameter.
The per-image inputs contain the camera-specific geometry embedding,
whereas each per-pixel input is a concatenation along the channel dimension of the RGB-based inputs, the pixel-specific geometry embedding, and the image indicator embedding.
As a result, the inputs are comprised of $2$ per-camera tokens with $D_{\textrm{cam}}$ dimensions, and $2\times\left(H\times W\right)$ per-pixel tokens of $\left(D_{\textrm{rgb}} + D_{\textrm{pix}} + D_{\textrm{ind}}\right)$ dimensions.
Finally, to ensure the two modes of input have the same channel dimension, we pad the smaller inputs with a learnable parameter.

\subsection{Featurizing Epipolar Cues}\label{sec:epipolar}
While the previous section uses on the geometric embedding to facilitate view triangulation, now we take one step further and exploit the given camera information to assist the search of correspondences between different images.

Correspondence estimation is paramount to multi-view geometry. 
The epipolar constraint is a fundamental constraint in stereo vision, which prescribes that a pair of corresponding points in the two images (projections of the 3D point) must lie on the corresponding epipolar lines, which are defined as the intersections between the image planes and the plane defined by the two camera centers and the 3D point (\cref{fig:geometry}).
In other words, a point in image 1 lying on epipolar line $l_1$ can only match to a point in image 2 that lies on the corresponding epipolar line $l_2$. Therefore, for a known camera pair one can compute the corresponding epipolar lines which can be used to restrict the search for point correspondences. This enables a drastically faster search while reducing the possibility of having outliers.

Similarly to camera information, the epipolar constraint is typically applied explicitly, \eg by restricting the correspondence search only along epipolar lines.
Instead, we provide the epipolar constraint directly as a part of the network inputs by tagging each pixel with its epipolar plane.
Note that each pixel is assigned to a single epipolar plane, apart from a special case which occurs only when the projection of the other camera (the epipole) falls inside the image, since all epipolar planes pass through the epipole; however, this degeneracy only potentially appears at a single pixel, and it is practically impossible for the epipole to align exactly with a pixel center, making this a non-issue.
Next, we consider two parameterizations of the epipolar plane.

The first option encodes the \emph{normal vector} of the epipolar plane, which can be easily computed as the normalized cross-product of $\vc_2-\vc_1$ and $\vr_{j,i}$, where $\vr_{j,i}$ is the ray direction in \eqref{eq:camera_rays}.
Formally, for pixel $i$ in image $j$, the normal vector, $\vn_{j,i}$, is:
\begin{align}
    \vv_{j,i} &= \left(\vc_2-\vc_1\right)\times \vr_{j,i} \\
    \vn_{j,i} &= \sign\left(\left[\vv_{j,i}\right]_{x}\right)\dfrac{\vv_{j,i}}{\lVert\vv_{j,i}\rVert+\epsilon},\label{eq:epipolar_normal}
\end{align}
where the $\left[\vv_{j,i}\right]_{x}$ is the $x$-coordinate of $\vv_{j,i}$ and the $\sign$ disambiguates the direction of the normal vectors (opposite normals denote the same plane).

The second option parameterizes the epipolar plane as a \emph{relative angle}, $\theta_{j,i}$, between the epipolar plane and an arbitrarily chosen reference epipolar plane, where the angles are scaled such that $\theta_{j,i}\in\left[-1,1\right]$ (\cref{fig:geometry}):
\begin{equation}
    \theta_{j,i} = 2\left(\frac{1}{\pi}\arccos\left(\dfrac{\vn_{j,i}^{\tr} \vn_{\textrm{ref}}}{\lVert\vn_{j,i}^{\tr} \vn_{\textrm{ref}}\rVert}\right)-0.5\right).
\end{equation}
We choose the reference epipolar plane, which is fixed for both frames, as the plane associated with a randomly chosen pixel from the first image.

Finally, for both parametrizations, the pixel-specific epipolar encodings, $\vn_{j,i}$ or $\theta_{j,i}$, are embedded into Fourier features and treated as `geometrical channels' (\cref{fig:input_matrix}), \ie concatenated to the per-pixel inputs along the channel dimension.

The epipolar embedding does not add new information compared to camera geometric embeddings described in \cref{sec:cameras}, but it provides an additional guidance for the network to more efficiently leverage correspondence.

\section{Experiments}

We evaluate our geometrical embeddings with the Perceiver IO model on the task of depth estimation from pairs of views, a central computer vision task.

\paragraph{Data.} We use the ScanNet~\cite{dai2017scannet} and DeMoN~\cite{ummenhofer2017demon} datasets for training and testing. 
For ScanNet we use the frame selection as provided by~\cite{kusupati2020normal}, which yields 94212 training pairs and 7517 test pairs.
The DeMoN dataset combines SUN3D~\cite{xiao2013sun3d}, RGBD-SLAM~\cite{sturm2012benchmark} and Scenes11~\cite{ummenhofer2017demon}.
It has a total of of 166,285 training image pairs from 50420 scenes and 288 test image pairs.
Both datasets contain invalid depth measurements, following the community common practice, we mask the depths out of $[0.1, 10]$ as invalid.

\paragraph{Implementation details.} We train our model with the commonly used \textsc{L1Log} loss~\cite{huynh2020guiding}, $\gL\left(d, d^\ast\right)=\lvert\log\left(d\right)-\log\left(d^\ast\right)\rvert$, where $d$ and $d^\ast$ are the predicted and ground truth depth values. 
Unless otherwise stated, we process images at $240\times320$ resolution.  
The raw RGB values are transformed to $64$-d (\ie $D_{\textrm{rgb}} = 64$) color features by a standard convolutional preprocessor described in Perceiver IO~\cite{jaegle2021perceiverIO}, which consists of 1-layer convolution with receptive field 7 and stride 2, followed by batch normalization, ReLU and stride-2 max-pooling, resulting in a feature grid of dimension $60 \times 80 \times 64$ for each image. 
Thess feature grids are combined with the geometric embeddings to form the inputs to the Perceiver IO model.
We use a small version of the original Perceiver IO architecture, which uses a $2048\times512$ matrix for the latent representation, 1 cross-attention for the input, followed by 8 self-attention layers and 1 cross-attention for the output, where the self-attention uses 8 heads and the cross-attention has only 1.
The output of the Perceiver IO model is two $60 \times 80$ depth maps. 
We upsample it by 4 to the original resolution with a Convex Upsampling module~\cite{teed2020raft} similar to Perceiver IO applied for optical flow estimation.

For the geometric embeddings, we consider relative camera pose \wrt the first camera. We set $K_{r}=K_{\textrm{matrix}}=K_p=10$, $K_o = 20$, the maximal sampling rate $\mu$ to is set to 60 for the $\vr_{j,i}$, $\vc_{j}$ and $\mP_{j}$, and to 120 for the epipolar cue.
These hyperparameters lead to the best evaluation in our empirical study.

We apply extensive augmentations, including random color jittering, which varies the brightness, contrast saturation and hue of the RGB inputs, as well as random cropping, rotation, and horizontal flipping.
We use the ADAM-W~\cite{kingma2014adam,loshchilov2017decoupled} optimizer with standard parameters $\beta_1=0.9$, $\beta_2=0.999$, and a cosine learning rate schedule without warmup, a weight decay of $1e{-5}$, a maximal learning rate of $2e{-4}$, and train for 250 epochs with a batch size of 64.

\subsection{Geometrical Embeddings}

We present results in a top-down matter, starting with the higher-level questions: are camera and epipolar geometrical embeddings useful? Are they complementary? We then trickle down and study finer-grained design decisions for each of these two families of geometrical embeddings. In this subsection all experiments are done on ScanNet. For statistical robustness, we train three models using different random seeds and report the median result.

\begin{table}[t!]
    \centering
    \begin{tabular}{ccccc}
    \toprule

camera & epipolar &
\multicolumn{3}{c}{training data proportion} \\
\cmidrule{3-5}
embedding & cue &
30\% & 50\% & 100\% \\\midrule
          &       &  0.2568  & 0.2423 & 0.2340  \\
\checkmark&          & \textbf{0.1350} & \textbf{0.1293} & 0.1234 \\
          &\checkmark& 0.2084 & 0.2018 & 0.1853 \\ 
\checkmark&\checkmark& 0.1371  & 0.1304 & \textbf{0.1204} \\
    \bottomrule
    \end{tabular}
    \caption{The effect of inputs on training efficiency and generalization (evaluated using absolute relative difference -- lower is better), using the best option for each mode.}
    \label{tab:ablations}
    \vspace{-3ex}
\end{table}

\paragraph{Coarse-grained analysis.}
We consider the best-performing options (according to the  fine-grained analysis in the next subsection) for camera and epipolar embeddings \cref{sec:cameras,sec:epipolar}.

\cref{tab:ablations} shows that, compared to using just standard pixel positional embeddings, any of the geometric embeddings contribute to substantial depth estimation accuracy improvement, with the camera embedding reducing the absolute relative difference almost by half. Interestingly, while the epipolar embedding itself does not provide sufficient information to perform triangulation, the epipolar embedding alone (row 3) can enhance the result as it provides additional guidance for correspondence estimation.

When provided with both the camera and epipolar embeddings (row 4), our model performs similarly as when using the camera embedding alone.
As the amount of training data increases however, the epipolar embedding seems to start contributing positively to the overall accuracy.

\paragraph{Fine-grained analysis.}
We now get down to more detailed analysis and compare the different options introduced in \cref{sec:cameras,sec:epipolar}.
First, we compare the two proposed camera parameterizations, namely using camera center and ray direction, $\vc_j $ and $\vr_{j,i}$, or directly using the projection matrix and the pixel positions, $\mP_j$ and $\vx_{j,i}$, as well as the two approaches for assembling this information into the input via geometrical channels or geometrical tokens.
As the upper part of \cref{tab:ablations_camera} shows, using camera center and ray direction has a consistent advantage regardless of the assembling method, likely thanks to its compactness.
At the same time, we observe that concatenating the geometric embedding channel-wise to the RGB inputs compares favorably with using the geometric embedding as a separate token. 
This is likely due to the fact that the concatenation provides a more direct association between the geometry and the pixel-wise RGB information.

Based on the best camera configuration, we evaluate the two options for the parameterization of the epipolar cue. 
As the lower part of \cref{tab:ablations_camera} shows, the angle parameterization slightly outperforms the normal parameterization, likely because the randomness in choosing the reference epipolar plane reduces the overfitting.

\begin{table}[tbhp]
\centering
\resizebox{0.48\textwidth}{!}{
\begin{tabular}{cccc}
\toprule
\makecell{camera \\parametrization} & \makecell{camera\\assembling} &
\makecell{epipolar\\parametrization} &\makecell{abs.\\rel.diff}\\\midrule
 $\vc_j $, $\vr_{j,i}$ & channel & -- & \textbf{0.1234} \\
 $\vc_j $, $\vr_{j,i}$ & token   & -- & 0.1249\\
 $\mP_j$, $\vx_{j,i}$  & channel & -- & 0.1345\\
 $\mP_j$, $\vx_{j,i}$  & token   & -- & 0.1805 \\\midrule
$\vc_j $, $\vr_{j,i}$ & channel  & $\vn_{j,i}$    &  0.1235\\
$\vc_j $, $\vr_{j,i}$ & channel  & $\theta_{j,i}$ & \textbf{0.1204} \\\bottomrule
\end{tabular}
}
\caption{Comparison between different parameterization options for camera and epipolar embeddings (using absolute relative difference).}
\label{tab:ablations_camera}\vspace{-3ex}
\end{table}

\begin{figure*}
    \centering
    \includegraphics[width=\textwidth, trim=0 1.5cm 0 0, clip]{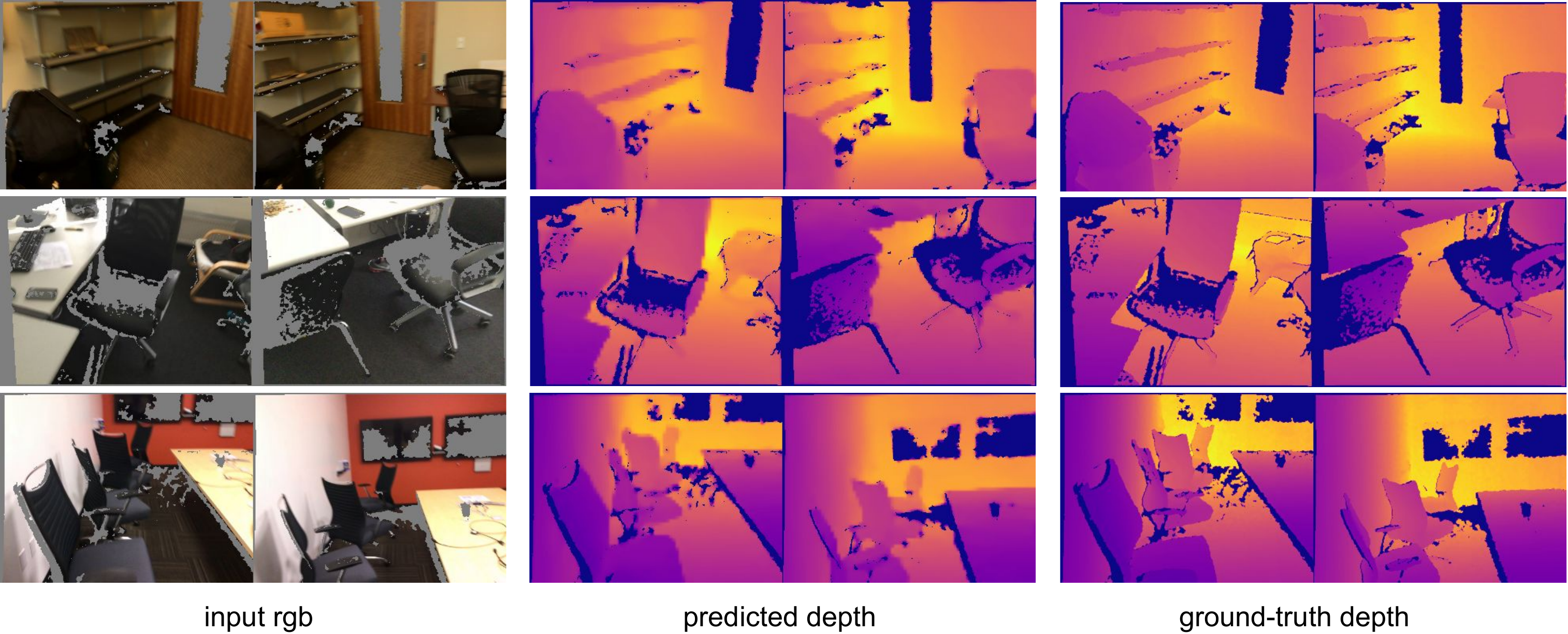} \\
    \begin{tabular*}{\textwidth}{ccc}
    ~~~~~~~~~~~~~~ Input image pairs &
    ~~~~~~~~~~~~~~~~~~~~~~~~~~~~~~~~~~ Model predictions &
    ~~~~~~~~~~~~~~~~~~~~~~~~~~~ Ground truth depth maps
    \end{tabular*}
    \caption{Examples of estimated depths using our best model on image pairs from ScanNet.
    Holes in the ground truth depth maps are masked out (shown in black).}
    \label{fig:qualitative}
\end{figure*}

\paragraph{Queries.}
We evaluate two types of queries.
As the first option, the queries take the same form as the inputs, using both the RGB features and the constructed geometric embeddings. Alternatively, we also experiment with discarding the RGB and querying using only the geometric embeddings.

\begin{figure}[t!]
    \centering
    \includegraphics[width=0.45\linewidth, height=0.4\linewidth]{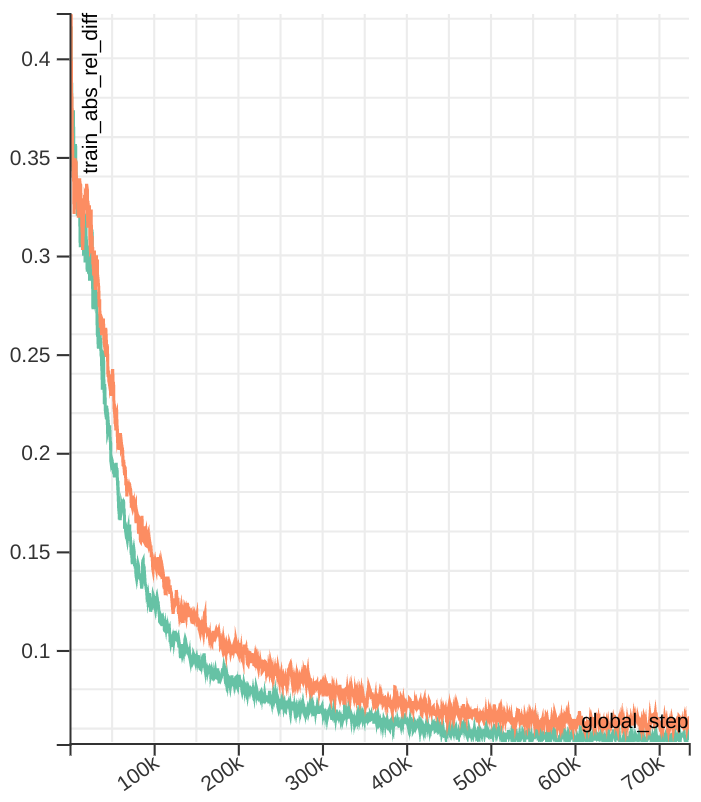}%
    \includegraphics[width=0.45\linewidth, height=0.4\linewidth]{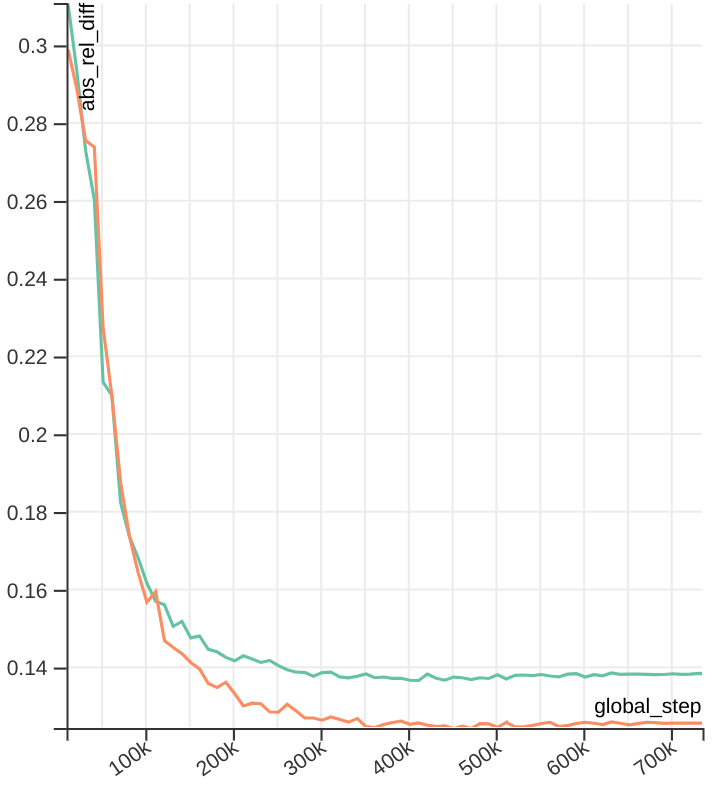}
    \caption{Queries with (green) and without (orange) RGB information. We show the training loss (left) and the validation curve (right). The change of the relative performance ranking between the two options in the validation curve suggests that overfitting to the RGB information.}
    \label{fig:query_curve}
    \vspace{-1ex}
\end{figure}

\begin{figure}
    \centering
    \includegraphics[width=\linewidth]{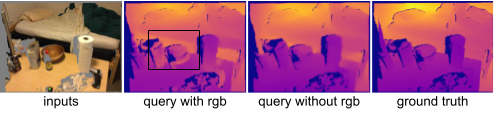}
    \caption{The effect of using RGB in the queries. Using RGB information in the query may cause artifacts due to overfitting the depth to texture.}
    \label{fig:rgb_artifact}\vspace{-3ex}
\end{figure}

We show the progression of training loss and validation error in \cref{fig:query_curve}.
We observe from the validation curve (right), that when including the RGB information (green) in the queries, the network initially learns slightly faster, but as the training progresses, this is outperformed by the queries that contain only the geometric-embedding.
On the other hand, the training loss of the RGB-included queries remains smaller than that of the RGB-excluded queries, suggesting that the RGB information eventually leads the network to overfit by overly attending to the texture information.
An example of such behavior is shown in \cref{fig:rgb_artifact}.

\subsection{Comparison with State-of-the-Art Methods}

We now compare our best model to the state-of-the-art on 4 different datasets: ScanNet, Sun3D, RGBD-SLAM and Scenes11. 
The results are shown in \cref{tab:stoa} and indicate that using geometrical embeddings with a very generic model matches and sometimes outperforms specialized state-of-the-art models. 
Note that the NAS model\cite{kusupati2020normal} uses additional ground truth normal information as supervision and enforces consistency between normals and depth.

We also evaluate the generalization ability of our method. As we demonstrate in the \cref{sec:app_generalization}, when testing on an unseen dataset of similar domain, our model performs on par with state-of-the-art methods trained specifically for that dataset, but
sees a performance drop under significant domain shift. This is somewhat expected, since unlike conventional plane-sweep methods, our model doesn't have the frames aligned externally but rather learns the alignment from input cues.

\begin{table}[t!]
    \centering\small
\resizebox{0.48\textwidth}{!}{
    \begin{tabular}{ccccc}
\toprule
dataset    &  methods  &   abs.rel. $\downarrow$  & rmse $\downarrow$ & \(\delta<1.25 \uparrow\) \\\midrule
\multirow{3}{*}{ScanNet} 
 & DPSNet\cite{im2019dpsnet} & 0.1258  & 0.3145 & - \\
 & NAS\cite{kusupati2020normal} & \textbf{0.1070}  &  \textbf{0.2807} & - \\
 & IIB (ours) & 0.1159 & \textbf{0.2807} & 0.9079 \\\midrule
\multirow{5}{*}{SUN3D} & DeMoN~\cite{ummenhofer2017demon} & 0.2137 & 2.4212 &  0.7332 \\
                       & DeepMVS\cite{huang2018deepmvs} & 0.2816 & 0.9436 & 0.5622\\
                       & DPSNet\cite{im2019dpsnet}  & 0.1469 & 0.4489 & 0.7812\\
                       & NAS\cite{kusupati2020normal}  & 0.1271 & 0.3775 & 0.8292\\
                       & IIB (ours) & \textbf{0.0985} & \textbf{0.2934} & \textbf{0.9018} \\\bottomrule
\multirow{5}{*}{\makecell{RGBD-\\SLAM}} & DeMoN~\cite{ummenhofer2017demon} & 0.1569 & 1.7798 &  0.8011\\
& DeepMVS\cite{huang2018deepmvs} & 0.2938  & 0.8684 &  0.5493\\
                       & DPSNet\cite{im2019dpsnet}  & 0.1508  & 0.6952 &  0.8041\\
                       & NAS\cite{kusupati2020normal}  & 0.1314  &  0.6190 & 0.8565\\
                       & IIB (ours) & \textbf{0.0951} & \textbf{0.5498} & \textbf{0.9065} \\\midrule
\multirow{5}{*}{\makecell{Scenes11}} & DeMoN~\cite{ummenhofer2017demon} & 0.5560 & 2.6034 &  0.4963\\
& DeepMVS\cite{huang2018deepmvs} & 0.2100  & 0.8909 &  0.6881\\
                       & DPSNet\cite{im2019dpsnet}  & 0.0500  & 0.4661 &  0.9614\\
                       & NAS\cite{kusupati2020normal}  & \textbf{0.0380}  &  \textbf{0.3710} &  \textbf{0.9754} \\
                       & IIB (ours) & 0.0556 & 0.5229 & 0.9631\\
                       \bottomrule

    \end{tabular}
}
\caption{Comparison with the state-of-the-art. Our method, here named IIB for \textit{Input-level Inductive Biases}, performs competitively with these more specialized methods, doing best on two of the four datasets and coming close in the other two.}
    \label{tab:stoa}
\end{table}

\begin{figure*}[ht]
    \centering
    \includegraphics[width=0.98\textwidth]{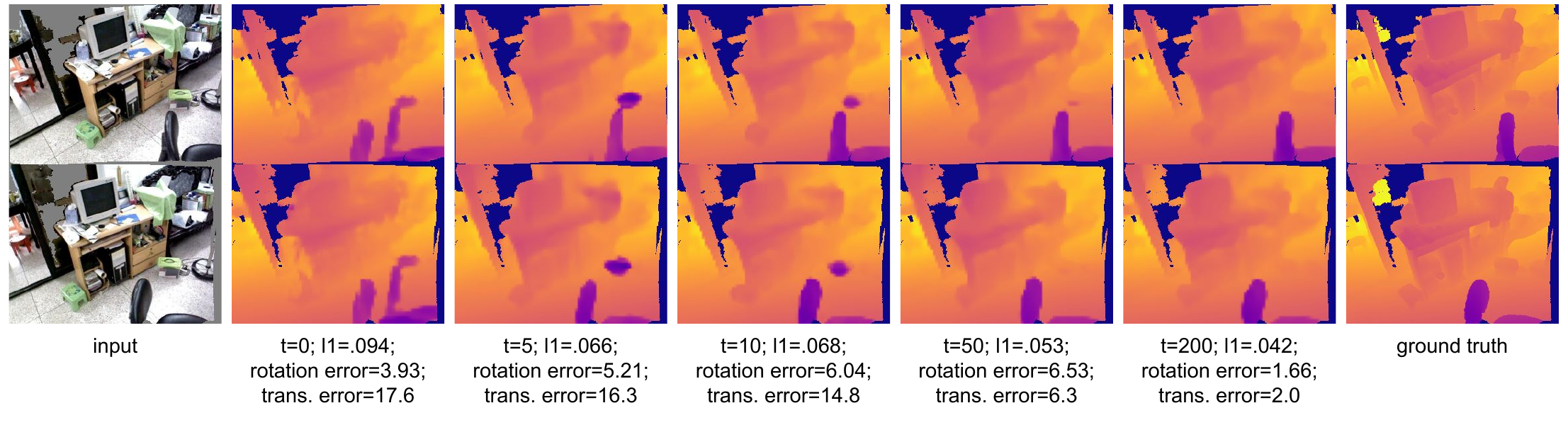}\vspace{-1ex}
    \caption{Progression of our iterative camera localization algorithm across 200 timesteps of optimization for the input image (left) and ground-truth depth that we are fitting to (right).  At each step we show the optimization timestep (t), l1 loss between the predicted depth and the ground truth depth, as well as the rotation and translation error between the estimated and ground truth cameras (which is not used in the optimization).}
    \label{fig:opt_progression}
\end{figure*}

\subsection{Camera Localization}
To what extent does our algorithm understand camera geometry, as opposed to simply memorizing depth~\cite{tatarchenko2019single}?  One way we can find out is by asking our algorithm to perform a useful task that it was never trained to perform.  It turns out that our network can actually localize cameras given 3D geometry, which is an important component of any SLAM system.  

We assume that we have a pair of images and ground truth depth maps for both.  We assume camera intrinsics are available, but the camera position and orientation are unknown.  We can randomly initialize the relative offset and orientation of the cameras, and then optimize them to minimize the \textsc{L1Log} distance between the predicted depth and ground truth depth.  Our underlying assumption is that errors in cameras will result in incorrect triangulation when the correspondences are correct.  Therefore, if the algorithm is doing correct 3D geometry, the error should be minimized when the relative camera positions are correct.  We give implementation details in \cref{sec:app_localization}.

\begin{table}[t!]
\centering
\resizebox{.48\textwidth}{!}{
\begin{tabular}{ccccc}\toprule
& \makecell{mean \\ rotation \\ error ($^{\circ}$)} &  \makecell{median \\ rotation \\ error ($^{\circ}$)} & \makecell{mean \\ translation \\ error (cm)} & \makecell{median \\ translation \\ error (cm)} \\
\hline
Identity & 9.11 & 7.61 & 17.7 & 12.7 \\
Rand. init & 9.18 & 7.68 & 17.8 &  12.8 \\
Optimized & 6.67 & 4.38 & 2.5 & 1.9 \\\bottomrule
\end{tabular}
}
\caption{Camera Localization performance on SUN3D. Lower is better.}
\label{tab:cam_localization}
\vspace{-1ex}
\end{table}

We evaluate on the SUN3D validation set. 
We treat the first camera as fixed at the origin, and evaluate the position of the second camera relative to it.  Following prior work on camera localization~\cite{sattler2018benchmarking}, we evaluate two metrics. First is translation error in cm, which is simply $\|c_{est}-c_{gt}\|_2$ where $c_{est}$ and $c_{gt}$ are the estimated and ground truth camera centers, respectively.  Second is rotation error in degrees, which is computed as $\mbox{arccos}((\mbox{trace}(R_{gt}^{-1}R_{est})-1)/2)$, where $R_{est}$ and $R_{gt}$ are the estimated and ground truth rotation matrices respectively.  This is the minimum rotation angle required to align both rotations. 

\cref{tab:cam_localization} shows our results.  We report both the mean and median across examples in the dataset.  We see non-trivial improvements in both metrics, localizing the cameras within a few centimeters and a few degrees of their true locations.  While we don't expect this to be competitive with SOTA SLAM systems (which typically integrate information across many more images), these results clearly show that the algorithm is using geometry as expected: depth error is minimized when the cameras are at the correct locations.  This may serve as a starting point for more flexible systems which don't rely on access to ground truth cameras.  

\cref{fig:opt_progression} shows how the depth map progresses throughout 200 steps of optimizing the camera.  We see that the initial depth is quite poor, with little definition on the desk, and a chair in the foreground which has been split in two, like double-vision in humans. These errors gradually resolve as the camera estimate gets better, with the algorithm able to correctly bind pixels across images using the geometry.  Interestingly, the translation error improves faster than the rotation error, suggesting that the algorithm may be using the camera centers more than the ray angles in order to perform triangulation.

\section{Discussion}
The 3D nature of space is a key aspect of our reality and should be incorporated as priors into our visual models. Most models for 3D reconstruction currently incorporate 3D priors by tailoring the architectures. In this paper we investigated an alternative inspired by advances in modeling with Transformers: we featurize these priors and feed them as inputs to the model. 
We show that this incurs no sacrifice in performance and in fact we obtain results that are competitive with leading models on several datasets. 
Our exploration in the space of geometry parameterization is non-exhaustive, more indicative priors may be derived to simplify the 3D reasoning.

Having geometric priors as inputs also opens up new possibilities: given a pre-trained frozen model and ground truth depth, one can finetune the geometrical inputs in case they are unknown, \eg for camera calibration or epipolar geometry estimation. 
Input-level inductive biases may also enable us to incorporate geometry into multimodal models, \eg, those that jointly process sound, touch or text. In such a setting, the type of architecture engineering that is appropriate for vision would no longer apply, whereas input-level biases still could. 

On the other hand, since our model needs to learn the 3D alignment, it expects the training and test data to have a similar distribution.
Moreover, since the baseline architecture operates on a compressed latent space, how the network solves the problem is potentially less interpretable because there are no explicit correspondence establishing steps.

\section{Acknowledgements}
We thank Yi Yang for providing advice regarding data processing and Jean-Baptiste Alayrac for his help with the training pipeline.
We are also grateful for Ankush Gupta for his insightful feedback on the paper.

{\small
\bibliographystyle{ieee_fullname}
\bibliography{egbib}
}

\appendix
\section{Generalization on Out-Of-Domain Test Data}\label{sec:app_generalization}
Our method generalizes to unseen datasets of a comparable domain. 
However, when testing on a significantly different domain, \eg trained on indoor scenes and testing on outdoor scenes, our framework will see a performance drop. 
Table~\ref{tab:generalisation} shows the performance of a model trained on ScanNet and evaluated on RGBD and SUN3D datasets.
SUN3D is similar to ScanNet, our method (trained on ScanNet only) performs reasonably well and on par with methods trained on SUN3D.
RGBD datasets contain many warehouse scenes where the depth range is significantly different from the one seen during training. 
Our model shows overfitting in this case.
This is because unlike conventional plane-sweep methods, where the network essentially computes a cost-volume from RGB features that are explicitly aligned, our method has to learn the alignment itself.
How to reduce such a gap in case of domain shift is a research direction for future work. 
Besides introducing more augmentation techniques, we think fine-tuning and scale-normalization are some promising directions to pursue.

\begin{table}[htpb]
\centering \small
\begin{tabular}{c@{~~~}c@{~~~}c@{~~~}c@{~~~}c@{~~~}c}\toprule
method & train & test  & abs.rel $\downarrow$ & rmse $\downarrow$ & $\delta < 1.25 \uparrow$ \\\midrule
IIB (ours) & ScanNet & SUN3D & 0.1291 & 0.3699 & 0.8298 \\
IIB (ours) & SUN3D & SUN3D & 0.0985 & 0.2934 & 0.9018 \\
NAS {[27]} & SUN3D & SUN3D & 0.1271 & 0.3775 & 0.8292 \\\midrule
IIB (ours) & ScanNet & RGBD & 0.2572 & 1.3102 & 0.5101 \\
IIB (ours) & RGBD & RGBD & 0.0951 & 0.5498 & 0.9065 \\
NAS {[27]} & RGBD & RGBD & 0.1314 & 0.6190 & 0.8565 \\\bottomrule
\end{tabular}
\caption{Generalisation performance.} \label{tab:generalisation}
\end{table}

\section{Camera localization: implementation details}\label{sec:app_localization}
In this section, we give implementation details for Section 4.3. Given a pair of images and ground truth depth maps, the goal is to infer the relative position and orientation of the two cameras using the network.  
We use a perceiver with $\vc_{j,i}, \vr_{j,i}$ for the camera embedding and $\vn_{j,i}$ for the epipolar constraint. 

Recall that cameras are parameterized with intrinsics $\mK$, and extrinsics $\mR$ and $\vt$.  We assume known $\mK$'s, and without loss of generality that the first camera is fixed as $\mR_1=\mI$ and $\vt_1=\vzero$. Thus, the extrinsics of the second camera $\mR_2$ and $\vt_2$ are the optimization variables, and these are parameterized as $\vt_2 \in \R^{3}$ and $\hat{\mR}_2 \in \R^{3\times 3}$.  To make sure that the extrinsic matrix is a rigid transformation, we compute the final rotation matrix as $\mR_2=\mU \mV$, where $\mU,\mS,\mV=\text{SVD}(\hat{\mR}_2)$, $\text{SVD}$ is singular value decomposition, $\mU$ and $\mV$ are orthonormal, and $\mS$ is diagonal. Therefore we are overparameterizing $\mR_2$ with 9 parameters but only 3 degrees of freedom, but empirically we find this gives good results.

Our optimization objective is \textsc{L1Log} loss on both images, plus regularization terms to encourage both the translation and rotation to be small.  Without regularization, we find that the optimization can get stuck in local minima with very large rotation and translation, which we don't expect will make sense to the network.  Specifically, we let $L_{rot}(\mR)=\mbox{arccos}((\mbox{trace}(\mR)-1)/2)$ in radians, and $L_{trans}(\vt)=\|\vt\|_{2}$.  Then the final loss can be written as:
$$L(\mR_2,\vt_2)=\mbox{\textsc{L1Log}}(\mR_2,\vt_2)+\lambda_{rot} L_{rot}(\mR_2) + \lambda_{trans} L_{trans}(\vt_2)$$
We set $\lambda_{rot}=1$ and $\lambda_{trans}=20$.  Note that during optimization, the translation coordinates use the default Sun3D scaling, which is $100\times$ smaller than what we report in our result table.   

We use the ADAM optimizer with a cosine learning rate schedule for 200 steps and an initial learning rate of $5e{-3}$.  We initialize $\vt_2=\epsilon_{trans}$ where $\epsilon_{trans}\in\R^{3}$ and $\hat{\mR}_2=\mI+\epsilon_{rot}$ where $\epsilon_{trans}\in\R^{3}$. Each element of both $\epsilon_{rot}$ and $\epsilon_{trans}$ is distributed as $\mathcal{N}(0,0.01)$.  We find that the network occasionally gets stuck in local optima, so we run with 5 different random initializations and take the solution with the best total loss $L$.  

\section{Qualitative Results}

\begin{figure*}
    \centering
    \def\qW{0.32}
    \begin{tabular}{ccc}
    \includegraphics[width=\qW\linewidth]{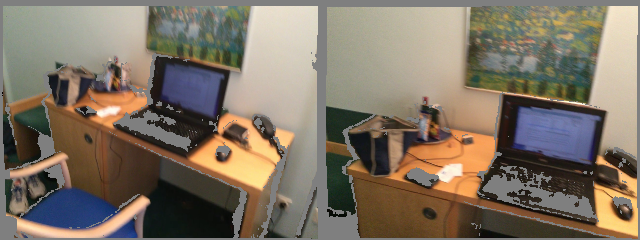} &
    \includegraphics[width=\qW\linewidth]{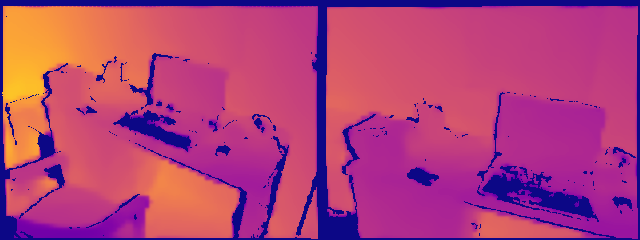} &
    \includegraphics[width=\qW\linewidth]{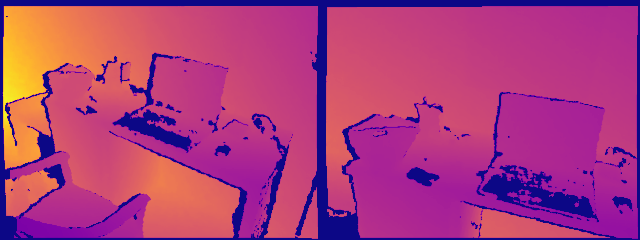} \\
    \includegraphics[width=\qW\linewidth]{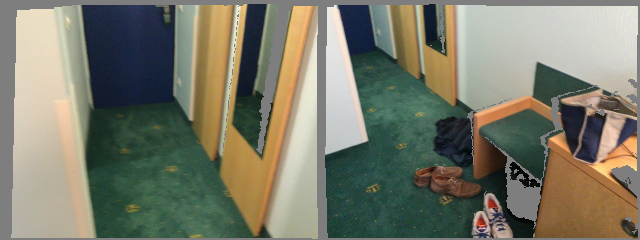} &
    \includegraphics[width=\qW\linewidth]{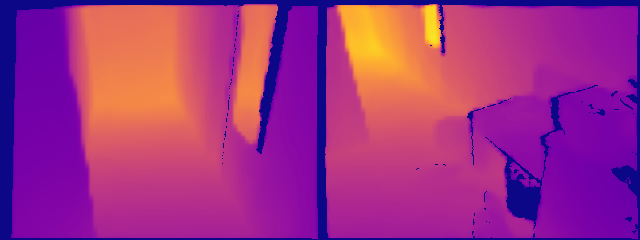} &
    \includegraphics[width=\qW\linewidth]{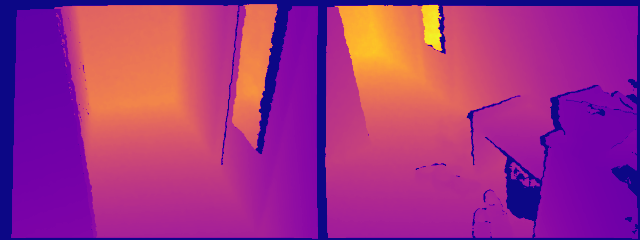} \\
    \includegraphics[width=\qW\linewidth]{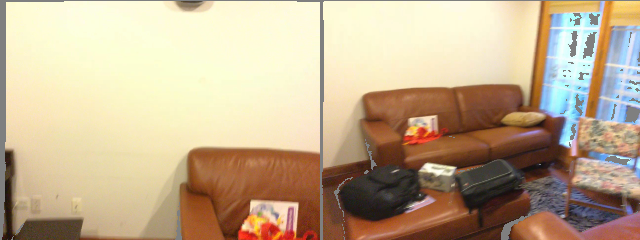} &
    \includegraphics[width=\qW\linewidth]{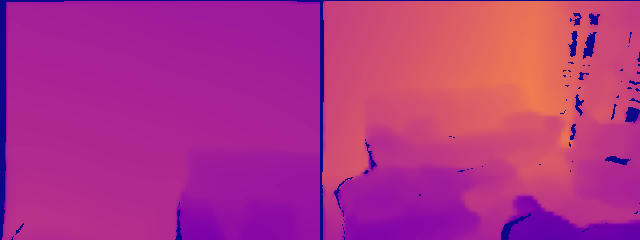} &
    \includegraphics[width=\qW\linewidth]{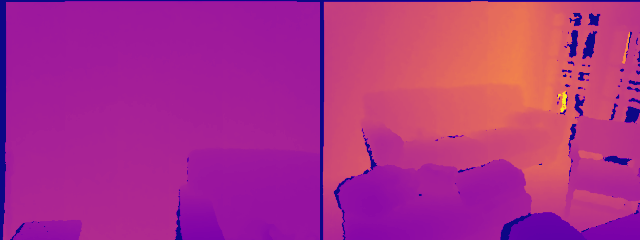} \\
    \includegraphics[width=\qW\linewidth]{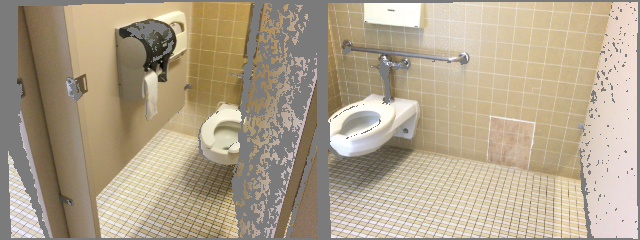} &
    \includegraphics[width=\qW\linewidth]{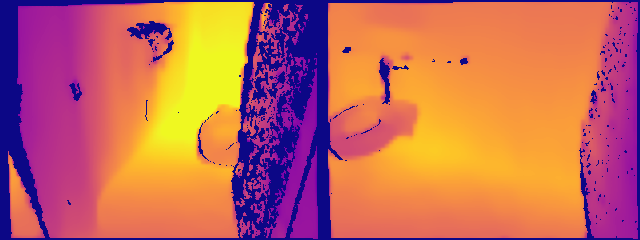} &
    \includegraphics[width=\qW\linewidth]{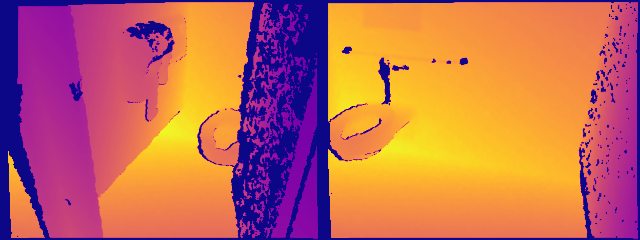} \\
    \includegraphics[width=\qW\linewidth]{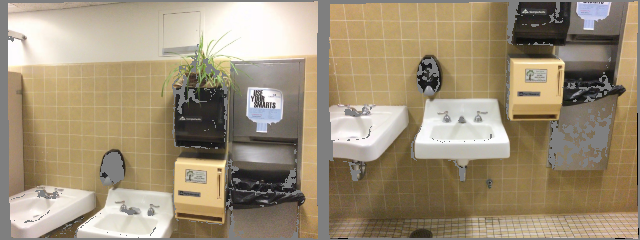} &
    \includegraphics[width=\qW\linewidth]{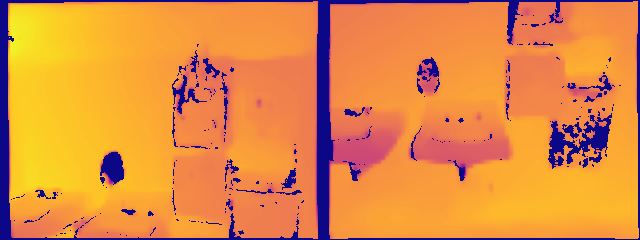} &
    \includegraphics[width=\qW\linewidth]{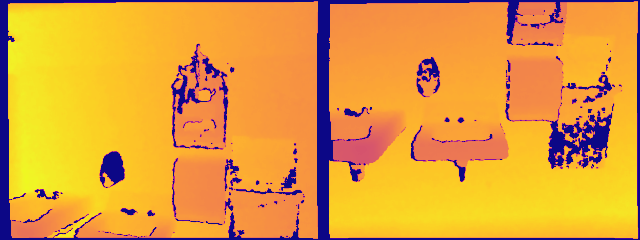} \\
    \includegraphics[width=\qW\linewidth]{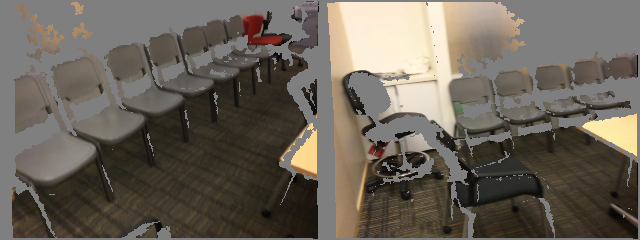} &
    \includegraphics[width=\qW\linewidth]{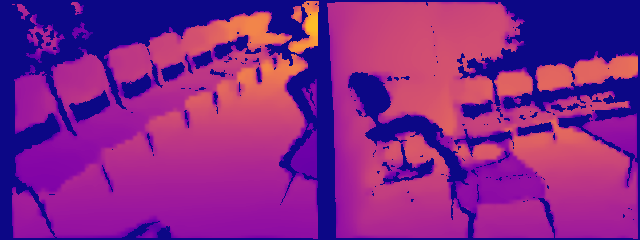} &
    \includegraphics[width=\qW\linewidth]{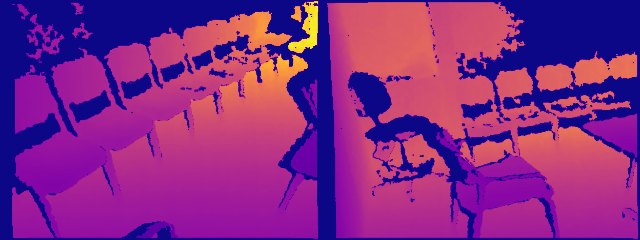} \\
    \includegraphics[width=\qW\linewidth]{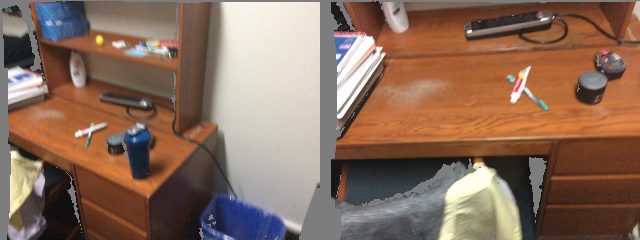} &
    \includegraphics[width=\qW\linewidth]{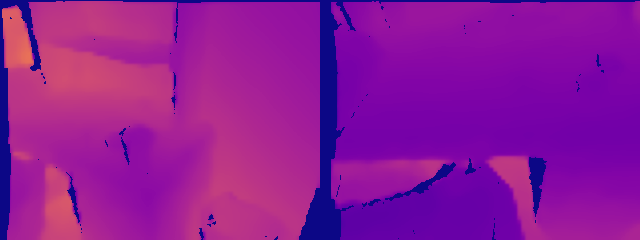} &
    \includegraphics[width=\qW\linewidth]{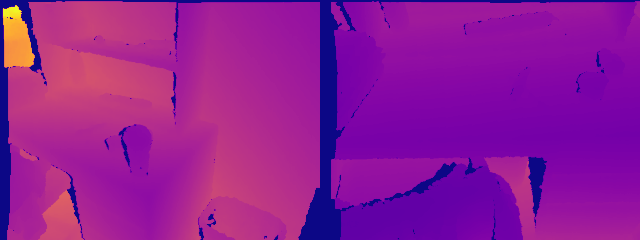} \\
    \includegraphics[width=\qW\linewidth]{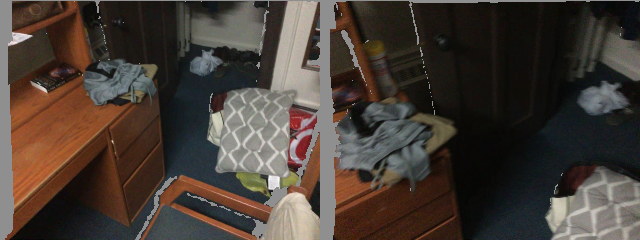} &
    \includegraphics[width=\qW\linewidth]{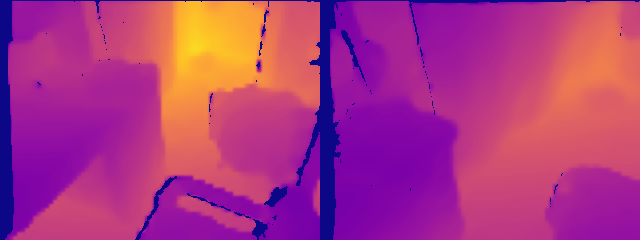} &
    \includegraphics[width=\qW\linewidth]{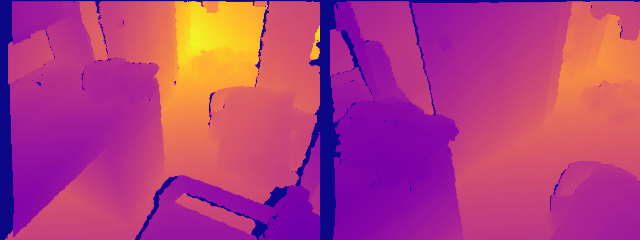} \\
    \includegraphics[width=\qW\linewidth]{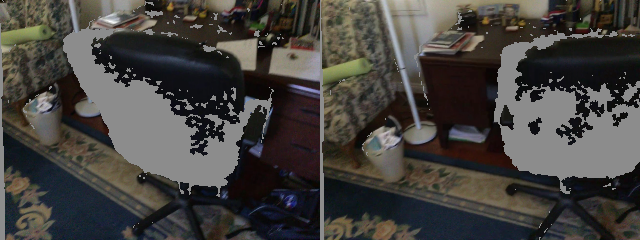} &
    \includegraphics[width=\qW\linewidth]{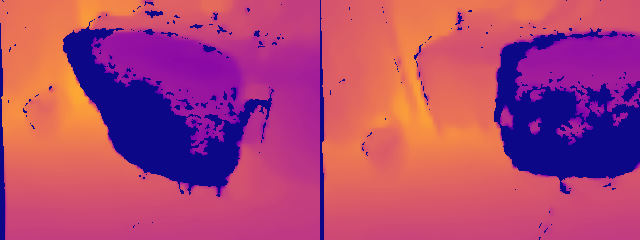} &
    \includegraphics[width=\qW\linewidth]{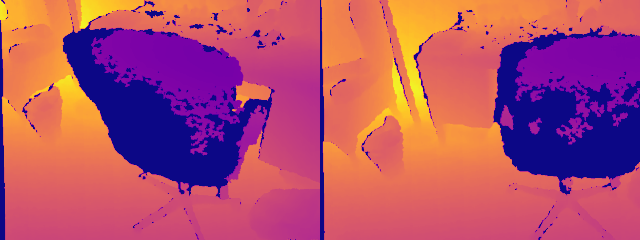} \\
    Input image pairs &
    Model predictions &
    Ground truth depth maps
    \end{tabular}
    \caption{Additional examples of estimated depths using our best model on image pairs from ScanNet.
    Holes in the ground truth depth maps are masked out (shown in black).}
    \label{fig:supp:qualitative}
\end{figure*}

Additional qualitative results are shown in Fig.~\ref{fig:supp:qualitative}.

\end{document}